\documentclass[letterpaper, 10 pt, journal, twoside]{ieeetran}
\usepackage{amsmath,amsfonts}
\usepackage{algorithmic}
\usepackage{algorithm}
\usepackage{array}
\usepackage[caption=false,font=normalsize,labelfont=sf,textfont=sf]{subfig}
\usepackage{stfloats}
\usepackage{url}
\usepackage{verbatim}
\usepackage{graphicx}
\usepackage{amsmath,amssymb,amsfonts}
\usepackage{algorithmic}
\usepackage{graphicx}
\usepackage{textcomp}
\usepackage{xcolor}
\usepackage[hidelinks,hypertexnames=false]{hyperref}
\usepackage{csvsimple}
\usepackage{cite}
\usepackage[absolute,overlay]{textpos}
\usepackage{blindtext}
\usepackage{threeparttable}
\usepackage{multirow}

\usepackage{pgfplots}
\usepackage{pgfplotstable}
\pgfplotsset{compat=1.16}

\usepackage{tikz}
\usetikzlibrary{calc}

\newcommand{\smat}[1]{\left[#1 \times\right]}

\definecolor{bblue}{HTML}{025497}
\definecolor{rred}{HTML}{C0504D}
\definecolor{ggreen}{HTML}{9BBB59}
\definecolor{ppurple}{HTML}{9F4C7C}
\definecolor{ggray}{HTML}{4D4D4D}

\begin{document}
\title{Radar-Based Odometry for Low-Speed Driving}

\author{Luis Diener, Jens Kalkkuhl, Markus Enzweiler
\thanks{This work has been submitted to the IEEE for possible publication. Copyright may be transferred without notice, after which this version may no longer be accessible.}
\thanks{This work is a result of the joint research project STADT:up (19A22006O). The project is supported by the German Federal Ministry for Economic Affairs and Climate Action, based on a decision of the German Parliament. The authors are solely responsible for the content of this publication.}
\thanks{L. Diener and J. Kalkkuhl are with Mercedes-Benz AG, Germany.}
\thanks{M. Enzweiler is with the Institute for Intelligent Systems, Esslingen University of Applied Sciences, Germany.}
}


\maketitle
\thispagestyle{plain}
\pagestyle{plain}

\begin{abstract}
We address automotive odometry for low-speed driving and parking, where centimeter-level accuracy is required due to tight spaces and nearby obstacles. Traditional methods using inertial-measurement units and wheel encoders require vehicle-specific calibration, making them costly for consumer-grade vehicles. To overcome this, we propose a radar-based simultaneous localization and mapping (SLAM) approach that fuses inertial and 4D radar measurements. Our approach tightly couples feature positions and Doppler velocities for accurate localization and robust data association. Key contributions include a tightly coupled radar-Doppler extended Kalman filter, multi-radar support and an information-based feature-pruning strategy. Experiments using both proprietary and public datasets demonstrate high-accuracy localization during low-speed driving.
\end{abstract}


\section{Introduction}
Accurate relative localization is critical for automated parking applications, where the vehicle executes low-speed maneuvers in complex environments. Unlike highway or urban driving, parking scenarios demand centimeter-level accuracy due to space constraints and the proximity to surrounding obstacles. The vehicle must follow the planned trajectory precisely and avoid obstacles while using the available space most effectively. Increasing the accuracy of relative localization and ego-motion estimation thus allows for lower safety margins, enabling the system to navigate into tighter parking spaces. Traditionally, automotive ego-motion estimation has relied on proprioceptive sensors such as inertial measurement units (IMUs), wheel encoders, and steering angle sensors \cite{Marco.2020}. These sensors provide high levels of reliability, availability, and update rates, making them capable of meeting automotive safety requirements. With the entry of automated driving functions into modern vehicles, additional sensors were required to perceive the vehicle's environment, e.g. cameras and radar sensors. As a result, these sensors are now broadly available in new vehicles.

\begin{figure}[t]
\includegraphics[width=\linewidth]{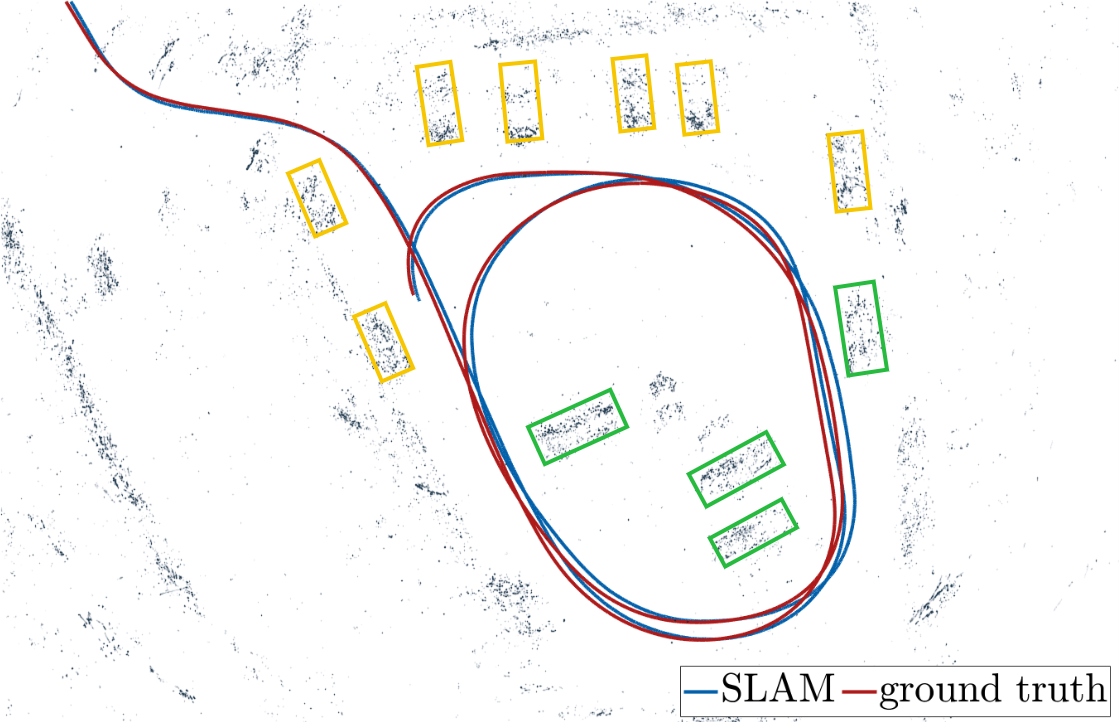}
\vspace{-0.6cm}
\caption{Our proposed radar SLAM algorithm on a parking space. Depicted are the solution of our proposed SLAM algorithm (blue), the ground-truth trajectory (red), and the radar point map.}
\label{fig:algo}
\vspace{-0.6cm}
\end{figure}

Current solutions for low-speed relative localization and ego-motion estimation rely on the fusion of IMU and wheel-encoder information \cite{Brunker.2019}. These algorithms can achieve precise relative localization when calibrated properly. However, this high accuracy can only be achieved for individually calibrated vehicles and not for an entire vehicle fleet, which makes the application to consumer-grade vehicles time consuming and costly. Simultaneous Localization and Mapping (SLAM) presents a promising solution to these challenges. SLAM introduces features into the estimation algorithm which can improve localization accuracy. This is especially relevant during parking maneuvers, where the same features can be observed over an extended period of time because of the low vehicle speeds. We focus on radar-based SLAM as current automotive-grade radar sensors offer four-dimensional (4D) measurements of the environment, i.e. position and Doppler velocity. We want to take full advantage of this 4D feature measurement and thus propose a tightly-coupled radar-based SLAM algorithm. Unlike existing approaches, we tightly couple the Doppler-velocity measurement with the feature states in the extended Kalman filter (EKF). This improves relative localization accuracy, especially at lower speeds, where features remain within the radar's field of view (FOV) for extended periods. Our main contributions are:
\begin{itemize}
\item A tightly-coupled radar-based SLAM algorithm using both Doppler and feature position measurements to estimate the vehicle's motion and the feature positions, improving accuracy.
\item An extension to multi-radar systems, where features are tracked and matched across all radar sensors. Individual features can thus provide more information to the filter even if they leave the FOV of an individual radar sensor.
\item A feature-pruning and maintenance strategy based on the provided information content to maximize the information flow to the filter.
\end{itemize}

Fig.~\ref{fig:algo} depicts our proposed SLAM algorithm with objects highlighted in the on-board camera images of the vehicle. The paper is structured as follows. Sec.~\ref{sec:related} provides an overview of radar odometry and SLAM methods. Sec.~\ref{sec:notation} briefly explains the notation used in the subsequent sections. The main contributions and the system are described in Sec.~\ref{sec:method}, while Sec.~\ref{sec:multi} covers the extension to multi-radar systems. This is followed by the experimental results in Sec.~\ref{sec:exp} and the conclusions in Sec.~\ref{sec:conclusions}.

\section{Related Work}
\label{sec:related}
With advances in automated driving applications consumer-grade vehicles are equipped with multiple 4D radar sensors that provide valuable information about the vehicle's ego motion. In this section, we summarize the current state of the art in radar odometry and radar-based SLAM and how these methods integrate IMU measurements. 

\subsection{Radar Odometry}
Kellner et al. \cite{Kellner.2013} introduced a well-known radar-only approach for ego-motion estimation, which computes the vehicle's velocity and yaw rate. Building on this work, Barjenbruch et al. \cite{Barjenbruch.2015} improved the methodology by incorporating the spatial movement of detections, using a gaussian mixture model. The method aligns with the approaches presented in later studies, that offer further improvements in handling uncertainties and correspondence matching \cite{Rapp.2017, Xu.2025, Haggag.2022}. Monaco and Brennan \cite{Monaco.2020} contributed by decoupling translational and rotational motions, using Doppler and spatial data respectively utilizing non-linear optimization to estimate rotational motion.
The integration of radar with other sensors has also been explored, such as in the work by Doer and Trommer \cite{Doer.2020} to fuse radar-Doppler measurements with inertial data. Diener et al. \cite{Diener.2024}, \cite{Diener.2025} incorporate online sensor and parameter estimation to improve the overall accuracy of ego-motion estimation for normal driving. In summary, the usage of radar-Doppler measurements allows for robust and accurate ego-motion estimation. In combination with spatial information, other sensors, and online calibration, researchers refined their methods further improving performance and robustness.

\subsection{SLAM}
SLAM differs from the previously discussed radar odometry approaches in that it introduces feature information into the estimation framework. SLAM simultaneously performs localization and feature estimation (mapping). The majority of recent developments in SLAM stem from visual or visual-inertial systems, which dominate the literature in this field. Nevertheless, the design choices and methodological differences within SLAM frameworks are mostly sensor independent. Among the most well-known SLAM systems are ORB-SLAM \cite{Campos.2021} and OKVIS \cite{Leutenegger.2015}, which use keyframe-based optimization in monocular and stereo camera setups. A critical design decision in SLAM concerns the parameterization of features, particularly in filter-based approaches where features are part of the system state. The conventional approach uses global 3D landmarks, that remain static in the world frame. However, to improve filter linearity and overall estimation consistency, several alternative representations have been proposed. For example, Civera et al. \cite{Civera.2008} separate depth and bearing components. Moreover, Bloesch et al. \cite{Bloesch.2017} introduce a minimal representation based on the unit sphere $S^2$, which reduces the dimensionality and improves the robustness of feature tracking. Although most SLAM research focuses on visual or visual-inertial systems \cite{Huang.2019}, radar-based localization is gaining interest due to its robustness in adverse environmental conditions. For instance, Michalczyk et al. \cite{Michalczyk.2022} propose a radar-inertial system that uses both radar distance and Doppler-velocity measurements to update the ego-motion state based on stochastic cloning. Other works focus on graph-based optimization \cite{Zhang.2023}, where Doppler information is used as a constraint \cite{Herraez.2024} and IMU data provides further refinements \cite{Herraez.2025}. Loop detection and closure is also applied to improve long-term accuracy, where the integration of Doppler measurements seems to have further positive effects on the overall accuracy as well \cite{Wang.2025}.

However, there is a lack of focus on high-accuracy, short-term localization for applications such as automated parking. Existing approaches rarely utilize multiple radar sensors, and cross-matching of features is not performed. Moreover, previous work directly uses the uncertain bearing measurements of the radar sensors for Doppler-velocity updates instead of coupling them to the estimated feature positions, leaving potential accuracy gains untapped. Our work aims to address these gaps by exploring tightly coupled multi-radar systems for precise, short-term localization and motion estimation. As we focus on short-term localization we do not perform loop closure. Moreover, we employ an EKF framework where we introduce both IMU states and features into the filter state.

\section{Notation}
\label{sec:notation}
We provide a brief overview of the employed notation. Three coordinate systems are used throughout the paper: the global world frame, ${G}$, the body-fixed coordinate frame, ${B}$, and the radar-fixed coordinate frame, ${R}$. The body-fixed coordinate frame has its origin at the IMU's position and is aligned with the vehicle axes (front, left, up). The global world frame is assumed to be flat, ignoring the Earth's curvature. Moreover, we ignore the rotation of the Earth since automotive-grade IMUs cannot measure it anyway.

Vectors are denoted as lower-case bold letters, e.g. $\mathbf{x}$, scalars as non-bold lower-case letters, e.g. $v$, and matrices as upper-case bold letters, e.g. $\textbf{F}$. Hamilton quaterions are used to represent attitude, e.g. ${}^{G}\mathbf{q}_{B}$, here describing the attitude of frame ${B}$ relative to frame ${G}$. The affiliated rotation matrix is ${}^{G}\mathbf{R}_{B}$, which is part of the special orthogonal group $SO(3)$ \cite{Sola.2017}. A similar notation is used for vectors, e.g. where ${}^{G}\mathbf{p}_{B}$ represents the position vector of frame ${B}$ to frame ${G}$, expressed in frame ${G}$. We define the following simplified notations for frequently used expressions:
\begin{itemize}
\item $\mathbf{p}_{B} := {}^{G}\mathbf{p}_{B}:$ position in the world frame,
\item $\mathbf{v}_{B} := {}^{B}\mathbf{v}_{B}:$ velocity in the body-fixed frame,
\item $\mathbf{q}_{B} := {}^{G}\mathbf{q}_{B}:$ attitude (body-fixed to world frame),
\item $\boldsymbol{\omega}_{B} := {}^{B}\boldsymbol{\omega}_{B}:$ angular rate in the body-fixed frame,
\item $\boldsymbol{a}_{B} := {}^{B}\boldsymbol{a}_{B}:$ acceleration in the body-fixed frame,
\end{itemize}

\section{System Description}
\label{sec:method}
\begin{figure*}[t]
\centering
\includegraphics[trim=40 0 20 0, clip, width=\textwidth]{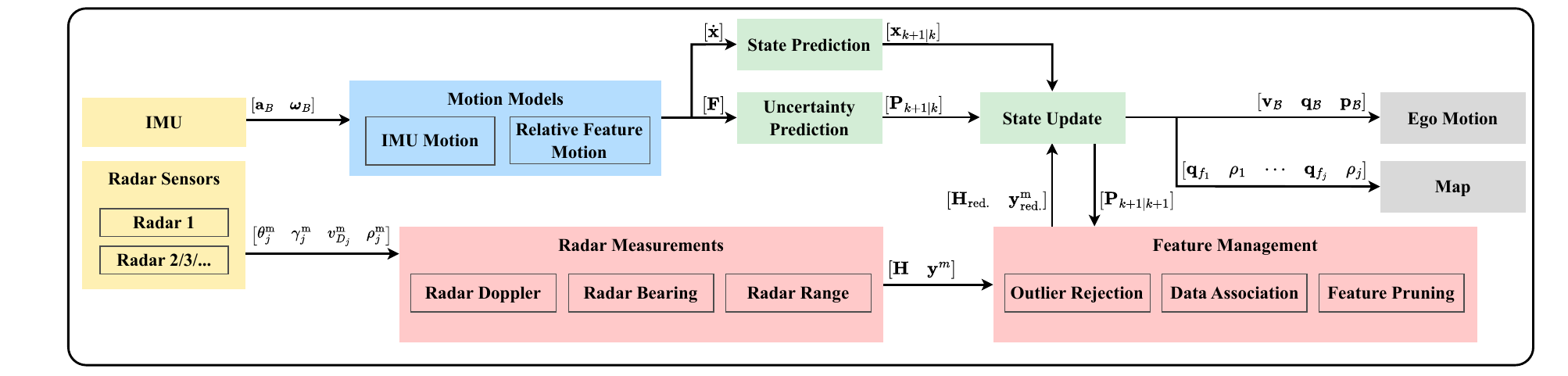}
\caption{Overview of the proposed method: numerical integration of both the IMU motion (Sec.~\ref{sec:motion}) and relative feature motion (Sec.~\ref{sec:feature}), tightly coupled with 4D radar measurements (Sec.~\ref{sec:radar}) and used for states estimation in an EKF framework (Sec.~\ref{sec:filter}) with prior feature management (Sec.~\ref{sec:management}).}
\label{fig:method}
\end{figure*}

We employ an EKF to estimate the vehicle's motion and the feature positions. We first describe the IMU motion model and the feature motion model, where we also define the IMU and features states. Subsequently, we introduce our set of radar measurement equations to tightly couple the Doppler-velocity measurement with SLAM features. This then allows us to explain the entire filter framework in detail with the necessary Jacobian matrices and prediction/update routines. The last section provides details on the feature management, i.e. outlier rejection, data association, and feature pruning. Fig.~\ref{fig:method} depicts an overview of our proposed framework.

\subsection{Motion Model}
\label{sec:motion}
The motion model describes the IMU's body-fixed velocity $\mathbf{v}_{B}$, its attitude $\mathbf{q}_{B}$ and the position $\mathbf{p}_{B}$ using inertial measurements as inputs. The derivatives of these states are expressed as \cite{Sola.2017}
\begin{align}
\label{eq:vmo_1}
\dot{\mathbf{v}}_{B} &= \mathbf{a}_{B}+\mathbf{R}(\mathbf{q}_{B})^\top{}^{G}\mathbf{g}-[\boldsymbol{\omega}_{B}\times] {\mathbf{v}}_{B}\\
\dot{\mathbf{q}}_{B} &= \frac{1}{2}\mathbf{q}_{B}\bullet\begin{bmatrix}
0\\ \boldsymbol{\omega}_{B} 
\end{bmatrix}\label{eq:vmo_2} \\
\dot{\mathbf{p}}_{B} &= \mathbf{R}(\mathbf{q}_{B}){\mathbf{v}}_{B} \label{eq:vmo_3}
\end{align}
with the gravity vector ${}^{G}\mathbf{g}$, the direction cosine matrix $\mathbf{R}(\mathbf{q}_{B})={}^{G}\mathbf{R}_{B}$, the acceleration $\mathbf{a}_{B}$, the angular rates $\boldsymbol{\omega}_{B}$, where the symbol $\{\bullet\}$ indicates quaternion multiplication, and where $[\boldsymbol{\omega}_{B}\times]$ is a skew-symmetric matrix.
The IMU state vector becomes
\begin{align}
\mathbf{x}_{B} &= \begin{bmatrix}
{\mathbf{v}}_{B}^\top & \mathbf{q}_{B}^\top&{\mathbf{p}}_{B}^\top\end{bmatrix}^\top
\end{align}
%
%
%
We extract the following Jacobians with respect to the IMU states:
\begin{align}
\dfrac{\partial\, \dot{\mathbf{v}}_{B}}{\partial\, \mathbf{v}_{B}}&=-[{\boldsymbol{\omega}_{B}}\times],\quad
\dfrac{\partial\, \dot{\mathbf{v}}_{B}}{\partial\, \mathbf{q}_{B}}={}^{B}\mathbf{R}_{G}[ {}^{G}\mathbf{g}\times]\\
\dfrac{\partial\, \dot{\mathbf{p}}_{B}}{\partial\, \mathbf{v}_{B}} &= {}^{G}\mathbf{R}_{B}, \quad \dfrac{\partial\, \dot{\mathbf{p}}_{B}}{\partial\, \mathbf{q}_{B}}=-[{}^{G}\mathbf{R}_{B}\mathbf{v}_{B}\times]
\end{align}

\subsection{Relative Feature Motion}
\label{sec:feature}
The stationary features exhibit a relative motion, that is entirely dependent on the vehicle's ego motion. Bloesch et al. \cite{Bloesch.2017} propose to parameterize features on the unit sphere $S^2$. This separates depth and bearing, which is especially relevant in vision-based approaches. However, this parameterization is also advantageous for radar sensors, since they measure both range and bearing independently. The bearing 
\begin{align}
\mathbf{q}_{f_j} := {}^{R}\mathbf{q}_{f_j}
\end{align}
of a feature $j$ relates to the feature's position in the radar frame as \cite{Bloesch.2017}:
\begin{align}
\mathbf{p}_{f_j} := {}^{R}\mathbf{p}_{f_j} = {}^{R}\mathbf{R}_{f_j}\mathbf{e}_1
\end{align}
The matrix ${}^{R}\mathbf{R}_{f_j}$ rotates the basis vector $\mathbf{e}_1$ from the feature's direction into the radar sensor frame, where $\mathbf{e}_{1/2/3}\in\mathbb{R}^3$ describe the basis vectors of an orthonormal coordinate system.
We also define the projection matrix:
\begin{align}
\mathbf{N}_\mathbf{q} := \mathbf{N}\left(\mathbf{q}_{f_j}\right) = {}^{R}\mathbf{R}_{f_j}\begin{bmatrix}\mathbf{e}_2&\mathbf{e}_3\end{bmatrix} \in \mathbb{R}^{3\times 2}
\end{align}
This matrix is orthogonal to the bearing vector $\mathbf{p}_{f_j}$ and is used to reduce the axis-angle representation to this orthogonal plane \cite{Jackson.2019}. 
The feature dynamics are given as \cite{Bloesch.2017}\cite{Jackson.2019}:
\begin{align}
\label{eq:feature1}
\dot{\mathbf{q}}_{f_j} &=-\mathbf{N}_\mathbf{q}^\top\left(\boldsymbol{\omega}_{R}+\rho_{j}\left[\mathbf{p}_{f_j}\times\right]\mathbf{v}_{R}\right)\\
\dot{\rho}_{j} &= -\left(\mathbf{p}_{f_j}\right)^\top\mathbf{v}_{R}
\label{eq:feature2}
\end{align}
where the radar-fixed velocity and angular rates are calculated using
\begin{align}
\mathbf{v}_{R} &= {}^{R}\mathbf{R}_{B} \left(\mathbf{v}_{B} + \smat{\boldsymbol{\omega}_{B}}{}^{R}\mathbf{p}_{B}\right) \\
\boldsymbol{\omega}_{R} &= {}^{R}\mathbf{R}_{B}\boldsymbol{\omega}_{B}
\end{align}
with the radar extrinsic alignment ${}^{R}\mathbf{R}_{B}$ and translation ${}^{R}\mathbf{p}_{B}$.
The dynamic feature state vector becomes:
\begin{align}
\mathbf{x}_f = \begin{bmatrix}
\mathbf{q}_{f_1}^\top&\rho_{1} & \mathbf{q}_{f_2}^\top&\rho_{2} & \cdots &\mathbf{q}_{f_j}^\top&\rho_{j}
\end{bmatrix}^\top
\end{align}
%
Bloesch et al. \cite{Bloesch.2016} provide the following general Jacobians \cite{Bloesch.2016} for the derivation:
\begin{align}
\dfrac{\partial\,\mathbf{p}_{f_j}}{\partial\, \mathbf{q}_{f_j}}&=-\left[\mathbf{p}_{f_j}\times\right]\mathbf{N}_\mathbf{q}\\
\dfrac{\partial\, \mathbf{N}_\mathbf{q}^\top\mathbf{r}}{\partial\, \mathbf{q}_{f_j}}&= \mathbf{N}_\mathbf{q}^\top\left[\mathbf{r}\times\right]\mathbf{N}_\mathbf{q},
\end{align}
where $\mathbf{r}$ is some vector, we can derive Jacobians for the feature dynamics. With respect to the feature states $\mathbf{x}_f$ we receive:
\begin{align}
\begin{split}
\dfrac{\partial\,\dot{\mathbf{q}}_{f_j}}{\partial\,\mathbf{q}_{f_j}} =& \mathbf{N}_\mathbf{q}^\top\left[\left(\boldsymbol{\omega}_{R}+[\mathbf{p}_{f_j}\times]\frac{\mathbf{v}_{R}}{\rho_{j}}\right)\times\right]\mathbf{N}_\mathbf{q}\\
&+ \mathbf{N}_\mathbf{q}^\top\frac{1}{\rho_j}[\mathbf{v}_{R}\times][\mathbf{p}_{f_j}\times]\mathbf{N}_\mathbf{q}
\end{split} \\
\dfrac{\partial\,\dot{\mathbf{q}}_{f_j}}{\partial\, \rho_{j}} &= \mathbf{N}_\mathbf{q}^\top[\mathbf{p}_{f_j}\times]\mathbf{v}_{R}\frac{1}{\rho_{j}^2}\\
\dfrac{\partial\,\dot{\rho}_{j}}{\partial\, \mathbf{q}_{f_j}} &= -\mathbf{v}_{R}^\top[\mathbf{p}_{f_j}\times]\mathbf{N}_\mathbf{q}
\end{align}
and with respect to the IMU states $\mathbf{x}_{B}$:
\begin{align}
\dfrac{\partial\,\dot{\mathbf{q}}_{f_j}}{\partial\, \mathbf{v}_{R}} &=-\mathbf{N}_\mathbf{q}^\top\frac{1}{\rho_{j}}[\mathbf{p}_{f_j}\times]\\
\dfrac{\partial\,\dot{\rho}_{j}}{\partial\, \mathbf{v}_{R}} &= -\mathbf{p}_{f_j}^\top
\end{align}
The depth dynamics provide the following Jacobians:
\begin{align}
\dfrac{\partial\,\dot{\rho}_{j}}{\partial\, \mathbf{q}_{f_j}} &= -\mathbf{v}_{R}^\top[\mathbf{p}_{f_j}\times]\mathbf{N}_\mathbf{q}\\
\dfrac{\partial\,\dot{\rho}_{j}}{\partial\, \mathbf{v}_{R}} &= -\mathbf{p}_{f_j}^\top
\end{align}

\subsection{Radar Measurements}
\label{sec:radar}
\begin{figure}[t]
\centering
\includegraphics[trim=20 10 20 10, clip, width=0.9\linewidth]{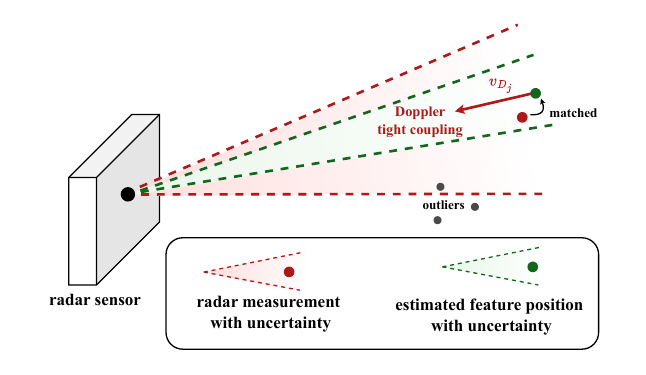}
\caption{The measured Doppler velocity (red) is applied to the estimated feature bearing (green), which exhibits lower uncertainty, thereby improving its contribution to the filter update.}
\label{fig:tight}
\end{figure}

We now introduce a novel set of measurement equations for radar SLAM applications. By adopting the unit sphere feature parameterization \cite{Bloesch.2017}, we can tightly couple Doppler measurements with tracked features. Thus, we use the estimated bearing of the feature that has lower uncertainty than the radar sensor's measured bearing vector. Fig.~\ref{fig:tight} depicts this key contribution and shows how it improves accuracy. At the same time, the separation of depth and bearing allows for a proper consideration of the distinct measurement noise characteristics. In the following derivations measured quantities are denoted with a superscript $y^\text{m}$.

The radar sensor measures the position of an object using azimuth $\theta_j^\text{m}$, elevation $\gamma^\text{m}_j$ and range $\rho_j^\text{m}$ measurements. Additionally, it measures the Doppler velocity ${v}^\text{m}_{D_j}$ of said object. This velocity denotes the relative, radial velocity between the radar sensor and the detected object. Assuming the detected object is stationary, the relationship between the radar's velocity $\mathbf{v}_{R}$ and the Doppler velocity ${v}^\text{m}_{D_j}$ becomes
\begin{align}
\label{eq:radar_vel}
{y}^\text{m}_{D_j}&=- v^\text{m}_{D_j}\\
 y_{D_j}&= {\mathbf{p}}_{f_j}^\top\mathbf{v}_{R}
\end{align}
where ${\mathbf{p}}_{f_j}^\top$ projects the velocity vector $\mathbf{v}_{R}$ onto the radial direction of the detected object $\{j\}$. Similarly, the object's bearing is measured as:
\begin{align}
{\mathbf{y}}^\text{m}_{q_j}&=\begin{bmatrix}
\cos\gamma^\text{m}_j\cos\theta^\text{m}_j & \cos\gamma^\text{m}_j\sin\theta^\text{m}_j & \sin\gamma^\text{m}_j
\end{bmatrix}^\top \\
\mathbf{y}_{q_j}&= {\mathbf{p}}_{f_j}
\end{align}
and its depth simply as:
\begin{align}
{y}^\text{m}_{\rho_j} &= {\rho}^\text{m}_j\\
{y}_{\rho_j} &= \rho_j
\end{align}
We receive the following measurement vector:
\begin{align}
\mathbf{y}^\text{m} = \begin{bmatrix}
{y}^\text{m}_{D_1} & {\mathbf{y}}_{q_1}^{m\,\top} & {y}^\text{m}_{\rho_1} & \cdots & {y}^\text{m}_{D_j} & {\mathbf{y}}_{q_j}^{m\,\top} & {y}^\text{m}_{\rho_j}
\end{bmatrix}^\top
\end{align}
For the three measurement functions, i.e. Doppler $y_D$, bearing $\mathbf{y}_{q_j}$, and depth $y_\rho$ we derive the Jacobians:
\begin{align}
\dfrac{\partial\, y_{D_j}}{\partial\, \mathbf{v}_{R}} &= {\mathbf{p}}_{sf_j}^{\,s,T},\quad
\dfrac{\partial\, y_{D_j}}{\partial\, \mathbf{q}_{f_j}} = \mathbf{v}_{R}^\top[\mathbf{p}_{f_j}\times]\mathbf{N}_\mathbf{q}\\
\dfrac{\partial\, \mathbf{y}_{q_j}}{\partial\,\mathbf{q}_{f_j}} &= [\mathbf{p}_{f_j}\times]\mathbf{N}_\mathbf{q}\\
\dfrac{\partial\, y_{\rho_j}}{\partial\,\rho_j} &= 1
\end{align}

\subsection{Filter Framework}
\label{sec:filter}
We first define the linearized continuous-time system that is used as the base for the EKF:
\begin{align}
\dot{\mathbf{x}} &= \mathbf{F}\,\mathbf{x} =\left[ \begin{array}{c|ccc}
\mathbf{F}_{B} & \multicolumn{3}{|c}{\mathbf{0}_{9\times 6+3j}}\\
\hline
\mathbf{F}_{B_1} &\mathbf{F}_{f_1} & &\\
\vdots & & \ddots &\\
\mathbf{F}_{B_j} & & &\mathbf{F}_{f_j}
\end{array}
\right]
\begin{bmatrix}
\mathbf{x}_{B}\\\mathbf{x}_f
\end{bmatrix} \\
\mathbf{y} &= \mathbf{H}\,\mathbf{x} = \left[
\begin{array}{c|ccc}
\mathbf{H}_{B_1} &  \mathbf{H}_{f_j} &  &\\
\vdots & &\ddots & \\
\mathbf{H}_{B_1}& & & \mathbf{H}_{f_j}
\end{array}
\right]
\begin{bmatrix}
\mathbf{x}_{B}\\\mathbf{x}_f
\end{bmatrix}
\end{align}
with the IMU state vector $\mathbf{x}_{B}$, the feature vector $\mathbf{x}_f$, its Jacobians:
\begin{align}
\mathbf{F}_{B} &= \left[
\begin{array}{ccc}
\frac{\partial \dot{\mathbf{v}}_{B}}{\partial\mathbf{v}_{B}} & \frac{\partial \dot{\mathbf{v}}_{B}}{\partial\mathbf{q}_{B}} & \mathbf{0}_{3}\\
\mathbf{0}_{3} & \mathbf{0}_{3} & \mathbf{0}_{3} \\
\frac{\partial \dot{\mathbf{p}}_{B}}{\partial\mathbf{v}_{B}} & \frac{\partial \dot{\mathbf{p}}_{B}}{\partial\mathbf{q}_{B}}& \mathbf{0}_{3}
\end{array}
\right]\\
\mathbf{F}_{B_j} &= \begin{bmatrix}
\frac{\partial\dot{\mathbf{q}}_{f_j}}{\partial \mathbf{v}_{B}} & \mathbf{0}_{2\times 3}& \mathbf{0}_{2\times 3}\\
\frac{\partial\dot{\rho}_{j}}{\partial \mathbf{v}_{B}}& \frac{\partial\dot{\rho}_{j}}{\partial \mathbf{q}_{f_j}} & \mathbf{0}_{1\times 3}
\end{bmatrix},\
\mathbf{F}_{f_j} = \begin{bmatrix}
\frac{\partial\dot{\mathbf{q}}_{f_j}}{\partial\mathbf{q}_{f_j}} & \frac{\partial\dot{\mathbf{q}}_{f_j}}{\partial \rho_{j}}
\end{bmatrix},
\end{align}
and the measurement matrices:
\begin{align}
\mathbf{H}_{B_j} &= \begin{bmatrix}
\frac{\partial\, y_{D_j}}{\partial\, \mathbf{v}_{B}} \\
\mathbf{0}_{4\times1}
\end{bmatrix},\ 
\mathbf{H}_{f_j} = \begin{bmatrix}
\frac{\partial\, y_{D_j}}{\partial\, \mathbf{q}_{f_j}} & 0\\
\frac{\partial\, \mathbf{y}_{q_j}}{\partial\,\mathbf{q}_{f_j}} & \mathbf{0}_{3\times1} \\
\mathbf{0}_{1\times2} & 1
\end{bmatrix}
\end{align}
We perform state estimation in discrete time within the EKF. 
The system's state $\mathbf{x}_{k|k}$ at $t=t_k$ is propagated using numerical integration of the continuous-time differential equations describing the IMU motion (Eqs.~\eqref{eq:vmo_1}-\eqref{eq:vmo_3}) and feature motion (Eqs.~\eqref{eq:feature1}-\eqref{eq:feature2}), resulting in the propagated state estimate $\mathbf{x}_{k+1|k}$. However, the covariance $P_{k|k}$ is propagated using the linearized discrete-time system:
\begin{align}
\dot{\mathbf{x}}_{k+1|k} &= \boldsymbol{\Phi}_k\,\mathbf{x}_{k|k} + \mathbf{w}_k \\
\mathbf{y}_k &= \mathbf{H}\,\mathbf{x}_{k|k} + \mathbf{v}_k
\end{align}
where $\mathbf{w}_k$ is the process noise with covariance matrix $\mathbf{Q}_k$, where $\mathbf{v}_k$ is the measurement noise with covariance matrix $\mathbf{R}_k$, and where the discrete-time state transition matrix is calculated as:
\begin{align}
\boldsymbol{\Phi}_k = \boldsymbol{\Phi}(t_{k+1},t_k) = \exp\left(\int_{t_k}^{t_{k+1}}\mathbf{F}(\tau)d\tau\right)
\end{align}
The state's covariance matrix $\mathbf{P}_{k|k}$ can then be propagated from $t_k$ to $t_{k+1}$:
\begin{align}
\mathbf{P}_{k+1|k} = \boldsymbol{\Phi}_k \mathbf{P}_{k|k} \boldsymbol{\Phi}_k^\top + \mathbf{Q}_k
\end{align}
For the time update within the EKF framework´, we first compute the Kalman gain:
\begin{align}
\mathbf{K}_k&= \mathbf{P}_{k+1|k}\mathbf{H}^\top \mathbf{S}_k^{-1},\\
\mathbf{S}_k &=\mathbf{H} \mathbf{P}_{k+1|k}\mathbf{H}^\top + \mathbf{R}_k,
\end{align}
using the innovation covariance $\mathbf{S}_k$.
Finally, we can update the current state and covariance:
\begin{align}
\mathbf{x}_{k+1|k+1} &= \mathbf{x}_{k+1|k} + \mathbf{K}_k \left(\mathbf{y}_k-{\mathbf{y}}^\text{m}_k\right),\\
\mathbf{P}_{k+1|k+1} &= \left(\mathbf{I}-\mathbf{K}_k\mathbf{H}\right)\mathbf{P}_{k+1|k}.
\end{align}

\subsection{Feature Management}
\label{sec:management}
The feature management handles three important tasks to ensure the health of the SLAM filter: \textbf{outlier rejection} ensures that only radar measurements of stationary targets are used to update the filter state and that clutter remains suppressed; \textbf{data association} describes the process of matching features of a particular measurement scan to the features within the filter; \textbf{feature pruning} is necessary to disregard features that left the FOV and to ensure information-rich features are added to the feature states.

The standard approach for data association and outlier rejection is nearest-neighbor matching using the Mahalanobis distance \cite{mahal.2018}. 
\begin{align}
\label{eq:mahal}
\mathbf{d}_M=\sqrt{\Delta\mathbf{y}^\top\mathbf{S}_{k}^{-1}\Delta\mathbf{y}},
\end{align}
with the innovation covariance $\mathbf{S}_{k}$.
This provides great results, if the sensor does not produce too much clutter and has a relatively low measurement uncertainty. There exist more sophisticated strategies such as optimizing the joint-compatibility of all matches \cite{Neira.2001} or maximizing the combined likelihood of the matches \cite{Zhang.2005}. However, we found that given our 4D measurements, the Mahalanobis distance based matching provides sufficiently accurate results and no additional benefit could be determined by any of the discussed. While the measurement uncertainty in both azimuth and elevation might produce multiple feature candidates, the Doppler-measurement accuracy usually steers the selection process towards a single detection. We thus use bearing, range and Doppler measurements for data association and outlier rejection using the Mahalanobis distance (Eq.~\eqref{eq:mahal}). This leads to the reduced measurement vector and matrix
\begin{align}
\mathbf{y}^\text{m}_\text{red.},\ \mathbf{H}_\text{red.}
\end{align}
Feature pruning is delicate, since on one hand we want to maintain features for as long as possible to increase performance, on the other hand the performance quickly degrades when the filter tracks an insufficient amount of features. Especially during parking we can expect features to remain within the vehicle's FOV for an extended period of time. Whereas, normal driving causes features to quickly leave the radar sensors' FOV. Some researchers develop heuristics based on feature age, successful tracking and overall filter health \cite{Bloesch.2017}. However, Zhang et al. \cite{Sen.2005} propose the concept of entropy for feature selection, quantifying the information increment $\Delta I_k$ each feature would provide in a state update:
\begin{align}
\label{eq:inf}
\Delta I_k &= I_{k+1}-I_k \\
I_k &= 0.5\ln\left((2\pi e)^n |P_k|\right),
\end{align}
where $P_k$ is the uncertainty matrix of the dynamic state vector at step $k$ and $n$ is the size of that state. We propose to use this framework two-fold; first, for feature pruning; second, for feature selection. For feature pruning we maintain a score for each feature that is increased according to Eq.~\eqref{eq:inf}, whenever a feature provided a successful update; and is decreased in a similar fashion whenever uncertainty of the dynamic states increases (filter prediction). Features with the lowest scores are then pruned in a regular interval. New features are chosen and initialized using the information (Eq. \ref{eq:inf}) they provide for the next update. Our feature pruning mechanism thus maximizes the information flow to the filter during both feature pruning and feature selection.

\section{Extension to Multi-Radar Systems}
\label{sec:multi}
We introduce an extension to our formalism that enables cross-feature matching between sensors, improving the accuracy and robustness of our algorithm. In contrast, state-of-the-art radar SLAM approaches, while potentially adaptable to multi-sensor setups, do not support cross-sensor feature association, limiting their performance. This is particularly relevant for parking maneuvers, where the vehicle executes sharp turns in confined spaces. There, features quickly leave a radar sensor's FOV and enter the FOV of an adjacent radar sensor. Cross-matching allows us to track features in these scenarios.

The features are still parameterized in the coordinate system of the sensor where they were first detected, causing no changes to the described algorithm when no cross-matching occurs. However, we now additionally transform features in-between radar sensors using their extrinsics. As an example, the transformation between sensor $R_1$ and sensor $R_2$ becomes
\begin{align}
{}^{R_2}\mathbf{p}_{f_j} = {}^{R_2}R_{R_1} {}^{R_1}\mathbf{p}_{f_j}\rho_{j}+{}^{R_2}R_{B}\left({}^{R_2}\mathbf{p}_{R_1}\right),
\end{align}
where ${}^{R_2}R_{R_1}$ is the coordinate transformation between the sensor coordinate systems, and where ${}^{R_2}\mathbf{p}_{R_1}$ is the shift between the coordinate systems.
This also affects the Doppler-velocity measurement:
\begin{align}
v_{D,j} = \frac{{}^{R_2}\mathbf{p}_{f_j}^\top }{|{}^{R_2}\mathbf{p}_{f_j}|}\mathbf{v}_{R_2}
\end{align}
Subsequently, the changed measurement equations impact the Jacobians of the filter update and thus also require minor adjustments. Using this approach we can apply the same data association and outlier rejection mechanism that we described in Sec.~\ref{sec:management}.

\section{Experimental Results}
\label{sec:exp}
\subsection{Experimental Setup}
We evaluate our algorithm using three datasets. One of those is a self-collected proprietary dataset and the other two are publicly available datasets: Hercules \cite{hercules.2025} and SNAIL radar set \cite{snail.2025}. Both public datasets use typical automotive-grade radar sensors (\textit{Continental ARS548}), high-precision IMUs, and they provide high-precision ground-truth measurements. Since both public datasets only offer one front-facing radar sensor, our multi-radar cross-matching functionality has no effect within these datasets. 

The vehicle of our proprietary dataset was equipped with four corner-radar sensors comparable in performance to the \textit{Continental ARS548}, featuring a detection range of up to 250~m and an angular accuracy of $0.5^{\circ}$. The automotive-grade IMU has a zero-bias stability of roughly 10~$^\circ$/h for the gyroscope and 5~m/s$^2$/h for the accelerometer. Additionally, the vehicle is equipped with a ground-truth measurement system achieving localization accuracy of approximately 2~cm. The extrinsic parameters of all sensors are known. End-of-line calibration is performed to obtain accurate alignment parameters of the radar sensors.

For our evaluation we allow for a maximum number of 50 features within the filter. The algorithm was implemented in an unoptimized research environment and executed on a standard desktop computer (Intel i7, 32 GB RAM).

\begin{figure}[t]
\centering
\includegraphics[width=0.49\linewidth]{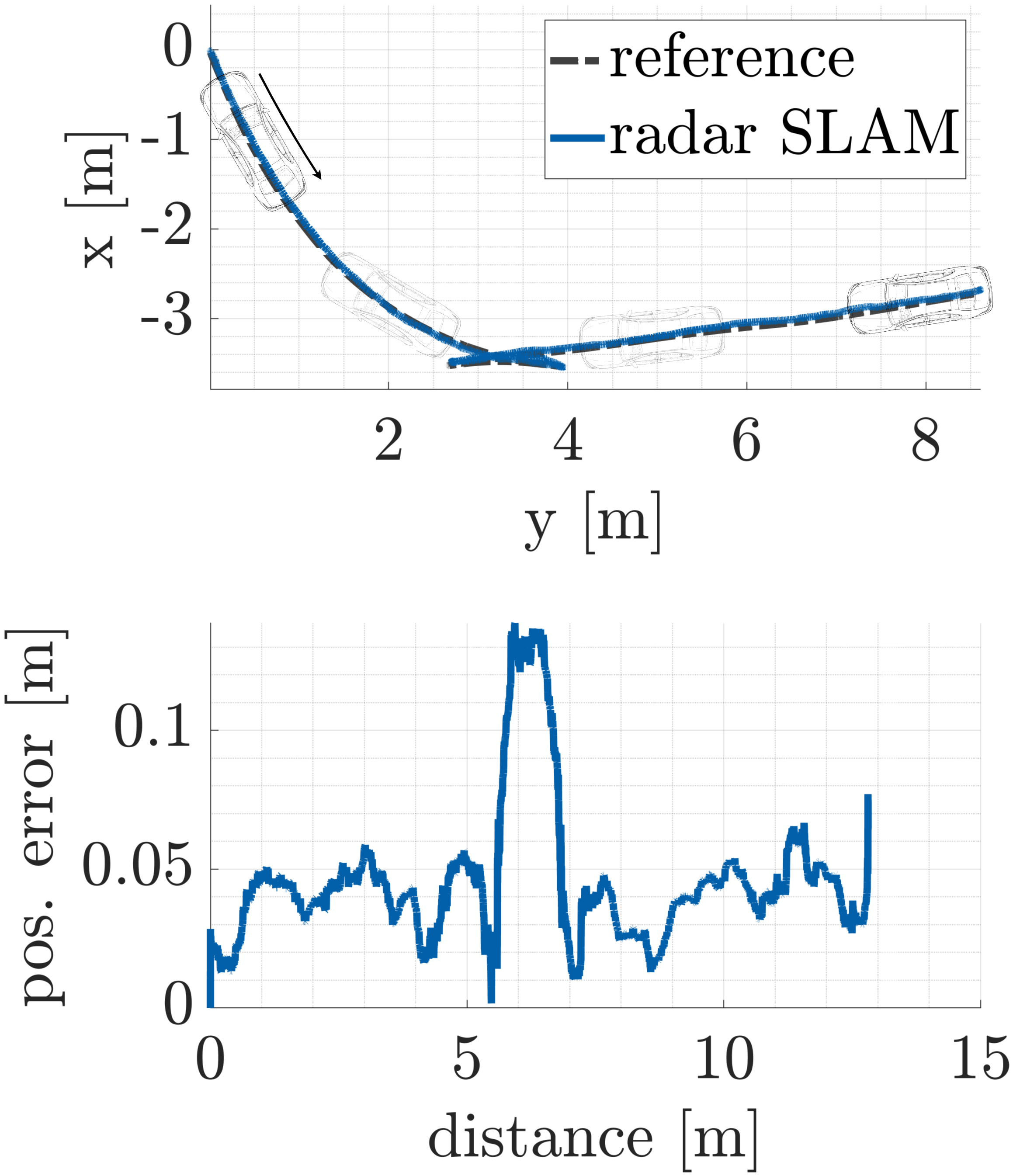}
\includegraphics[width=0.49\linewidth]{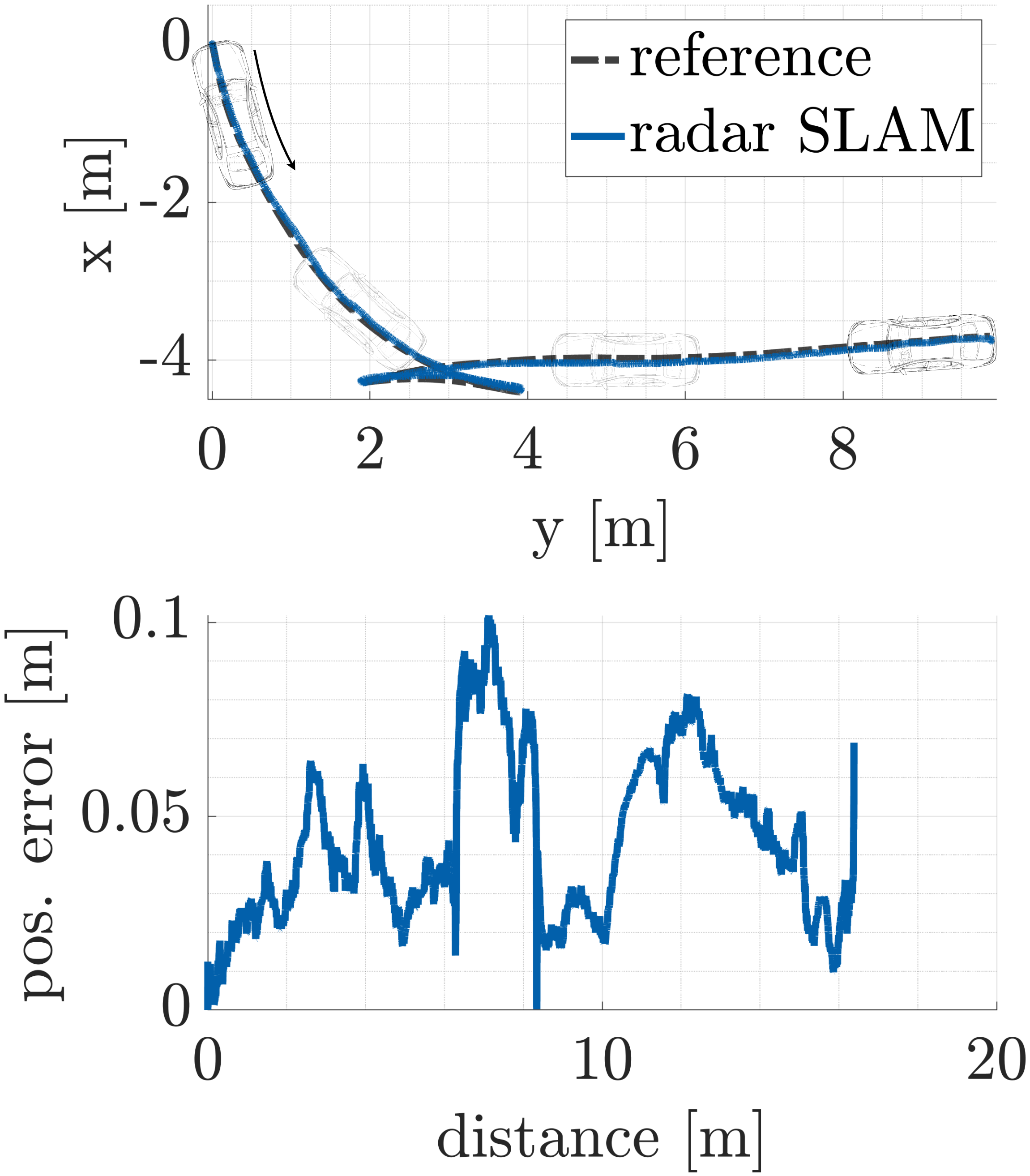}
\caption{Localization for two perpendicular parking maneuvers. The upper plots show the trajectories, while the plots below depict the error between the reference trajectory and the SLAM trajectory.}
\label{fig:plot_03}
\end{figure}

\subsection{Proprietary Datasets}
Our proprietary dataset offers 54 distinct parking maneuvers, each beginning and ending with the vehicle at a standstill. Two primary types of maneuvers are included: parallel parking and perpendicular parking (majority). Each individual parking maneuver has a length between 10-20 m, with the vehicle starting in front of the parking spot and then executing the parking maneuver. Fig.~\ref{fig:plot_03} depicts two examples of perpendicular parking maneuvers. In both examples our SLAM algorithm estimates the vehicle's end position within 8 cm of the true value. 
To properly evaluate the performance of our algorithm, we consider two metrics: the end-pose error and the trajectory error. The end-pose error represents the pose error at the end of each parking maneuver and shows how much error is accumulated within each individual parking maneuver. The trajectory error depicts the root-mean-square error (RMSe) over each individual parking maneuver's full trajectory eliminating end-pose bias.

\renewcommand{\arraystretch}{1.3}
\begin{table}[b]
\centering
\begin{threeparttable}
\caption{Evaluation of localization accuracy for perpendicular parking maneuvers.}
\label{tab:eval}
\begin{tabular}{l|ccc}
\hline
& {\textbf{63rd percentile}} & \textbf{95th percentile} & \textbf{maximum}\\
\hline
end pose error & {{0.13 m}} & {{0.24 m}} & {0.28 m}\\
trajectory error & {{0.10 m}} & {{0.17 m}} & {0.28 m}\\
\end{tabular}
\begin{tablenotes}
\item Note: The values for the 63rd and the best 95th percentile indicate that 63\% or 95\% of all samples have an error of less than or equal to the value provided in the table. Depicted are the both the error at the end of the maneuver and the error over the whole trajectory.
\end{tablenotes}
\end{threeparttable}
\end{table}

We perform a statistical evaluation over all 54 parking maneuvers.
Table \ref{tab:eval} depicts the results both for both the end-pose error and the trajectory error. We evaluate these errors within their 63rd percentile, 95th percentile and maximum values.
In 63\% of the test cases (performance threshold) a localization accuracy of 10 cm is achieved with the end position being estimated within 13 cm. Our algorithm achieves a consistent localization performance (95th percentile) of 17 cm and a maximum recorded error of 28 cm. These results are comparable to calibrated wheel-odometry-based algorithms \cite{Brunker.2019}. However, the advantage of our radar-based approach is its robustness against wheel slip and a lower vehicle-calibration effort, making the algorithm suitable for consumer-grade applications, where individual vehicle calibration is unfeasible.

\subsection{Ablation Study}

\renewcommand{\arraystretch}{1.1}
\begin{table}[t]
\centering
\begin{threeparttable}
\caption{Ablation Study on End Pose Error of Perpendicular Parking Maneuvers.}
\label{tab:ablation}
\begin{tabular}{l|wc{1.7cm}wc{1.7cm}c}
\hline
& \textbf{63rd percentile} & \textbf{95th percentile} & \textbf{max.} \\
\hline
w/o Doppler velocity & \textbf{0.20 m} & \textbf{{0.37 m}}& \textbf{{0.58 m}}\\
w/o cross matching & \underline{0.17 m} & 0.28 m & 0.36 m\\
1/2 feature count (25) & 0.13 m & {0.24 m} & {0.28 m}\\
1/4 feature count (12) & 0.16 m & \underline{0.29 m} & \underline{0.48 m}\\
\hline
\textbf{full solution} & 0.13 m & {0.24 m} & {0.28 m}
\end{tabular}
\begin{tablenotes}
\item Note: Values in \textbf{bold} indicate the greatest impacts and \underline{underlined} values indicate the second biggest changes.
\end{tablenotes}
\end{threeparttable}
\end{table}

We also want to assess the individual impact of our key contributions on the overall performance. Specifically, we evaluate the tightly-coupled Doppler-velocity measurement update; cross-sensor feature tracking; and the sensitivity of the system to the maximum number of features allowed within the filter.
The results are summarized in Tab. \ref{tab:ablation}. We observe that reducing the number of tracked features has minimal impact on performance until the number is lowered to 12 features, at which point the localization accuracy degrades noticeably. Among the evaluated components, the tightly-coupled velocity update yields the largest improvement, with its removal leading to the largest performance drop overall. Additionally, cross-sensor feature matching shows a clear benefit, confirming that successful tracking across sensors contributes meaningfully to the overall accuracy.
These results demonstrate that our tightly-coupled velocity update and cross-sensor feature tracking are key contributors to the localization accuracy of our algorithm.

\renewcommand{\arraystretch}{1.3}
\begin{table*}[b]
\centering
\resizebox{\textwidth}{!}{
\begin{threeparttable}
\caption{Benchmark snail Dataset.}
\label{tab:snail}
\begin{tabular}{l|ccc|ccc|ccc|ccc|ccc}
\hline
Sequence & \multicolumn{3}{c|}{\textbf{113/1 (Parking)}} & \multicolumn{3}{c|}{\textbf{113/3}}& \multicolumn{3}{c|}{\textbf{115/2}}& \multicolumn{3}{c|}{\textbf{123/2}}& \multicolumn{3}{c}{\textbf{123/3}}\\
\hline
Evaluation Metric & RPE [m] & RRE [$^\circ$] & APE [m] & RPE [m] & RRE [$^\circ$] & APE [m] & RPE [m] & RRE [$^\circ$] & APE [m] & RPE [m] & RRE [$^\circ$] & APE [m] & RPE [m] & RRE [$^\circ$] & APE [m] \\
\hline
Radar-ICP \cite{Herraez.2024} & 0.238 & 0.156 & 18.2 & 0.120 & 0.174 & 3.9 & 0.229 & 0.131 & 31.6 & 0.252 & 0.112 & 37.5 & 0.221 & 0.151 & 7.9 \\
RIV-SLAM* \cite{Wang.2024} & 0.213 & 0.142 & 30.2 & \underline{0.113} & 0.171 & 4.1 & \underline{0.219} & 0.128 & 33.1 & \underline{0.224} & 0.101 & 35.5 & 0.201 & 0.140 & 6.1 \\
RaI-SLAM* \cite{Herraez.2025} & 0.242 & 0.147 & 4.7 & 0.115 & 0.164 & \underline{3.4} & 0.220 & 0.117 & \underline{7.9} & 0.256 & 0.104 & \underline{8.4} & 0.219 & 0.139 & \underline{3.5} \\
Doppler-SLAM* \cite{Wang.2025} & \underline{0.150} & \underline{0.111} & \underline{1.53} & \textbf{0.082} & \underline{0.160} & \textbf{0.32} & \textbf{0.174} & \textbf{0.098} & \textbf{5.65} & \textbf{0.198} & \textbf{0.095} & \textbf{5.81} & \underline{0.156} & \underline{0.116} & \textbf{1.556}\\
\hline
\textbf{Our Approach} & \textbf{0.027} & \textbf{0.089} & \textbf{0.53} & 0.240 & \textbf{0.090} & 48.91 & 0.263 & \underline{0.107} & 34.11 & 0.269 & \underline{0.097} & 37.29 & \textbf{0.098} & \textbf{0.105} & 8.87
\end{tabular}
\begin{tablenotes}
\item Note: The best result is depicted in \textbf{bold} and the second best solution is \underline{underlined}. We use the same evaluation metrics as provided by Wang et al. \cite{Wang.2025} with a sampling rate of 2 Hz.\\
* These algorithms employ loop detection and closure.
\end{tablenotes}
\end{threeparttable}
}
\vspace*{-5pt}

\centering
\resizebox{\textwidth}{!}{
\begin{threeparttable}
\caption{Benchmark Hercules Dataset.}
\label{tab:hercules}
\begin{tabular}{l|ccc|ccc|ccc|ccc|ccc}
\hline
Sequence & \multicolumn{3}{c|}{\textbf{Mountain Day 1}} & \multicolumn{3}{c|}{\textbf{Library Day 1}}& \multicolumn{3}{c|}{\textbf{Sports Complex Day 1}}& \multicolumn{3}{c|}{\textbf{Parking Lot Night 3}}& \multicolumn{3}{c}{\textbf{Street Day 1}}\\
\hline
Evaluation Metric & RPE [m] & RRE [$^\circ$] & APE [m] & RPE [m] & RRE [$^\circ$] & APE [m] & RPE [m] & RRE [$^\circ$] & APE [m] & RPE [m] & RRE [$^\circ$] & APE [m] & RPE [m] & RRE [$^\circ$] & APE [m] \\
\hline
Radar-ICP \cite{Herraez.2024} & \underline{0.405} & 0.309 & 118.6 & 0.288 & 0.338 & 10.23 & 0.286 & 0.381 & 7.13 & 0.263 & 0.521 & 3.41 & \underline{0.070} & 0.890 & 11.66 \\
RIV-SLAM* \cite{Wang.2024} & 0.480 & 0.202 & 98.33 & \textbf{0.030} & \textbf{0.069} & \textbf{4.18} & 0.363 & 0.167 & 12.57 & 0.198 & 0.165 & 2.38 & 0.136 & 0.105 & 10.74 \\
Doppler-SLAM* \cite{Wang.2025} & \textbf{0.306} & \textbf{0.158} & \underline{43.05} & \underline{0.124} & \underline{0.128} & 13.24 & \textbf{0.143} & \textbf{0.127} & \textbf{2.72} & \underline{0.160} & \underline{0.151} & \textbf{0.717} & 0.089 & \underline{0.095} & \underline{7.45} \\
\hline
\textbf{Our Approach} & 0.409 & \underline{0.193} & \textbf{28.78} & 0.143 & 0.142 & \underline{6.42} & \underline{0.163} & \underline{0.164} & \underline{4.67} & \textbf{0.057} & \textbf{0.186} & \underline{3.12} & \textbf{0.049} & \textbf{0.062} & \textbf{5.19}
\end{tabular}
\begin{tablenotes}
\item Note: The best result is depicted in \textbf{bold} and the second best solution is \underline{underlined}. We use the same evaluation metrics as provided by Wang et al. \cite{Wang.2025} with a sampling rate of 2 Hz.\\
* These algorithms employ loop detection and closure.
\end{tablenotes}
\end{threeparttable}
}
\end{table*}

\subsection{Public Datasets}
To enable a comparison with existing approaches, we evaluate our method on publicly available datasets. Specifically, we use the Hercules dataset \cite{hercules.2025} and the SNAIL radar set \cite{snail.2025}, as both provide automotive-grade radar sensors, ground-truth trajectories, and inertial measurements. Moreover, both datasets include low-speed sequences. We benchmark our approach against recent state-of-the-art radar-based SLAM algorithms, namely Radar-ICP (radar-only odometry) \cite{Herraez.2024}, RIV-SLAM (radar-inertial graph-based SLAM) \cite{Wang.2024}, RaI-SLAM (Doppler-aided graph optimization) \cite{Herraez.2025}, and Doppler-SLAM (Doppler-filter front end with a graph-optimization back end) \cite{Wang.2025}.
Evaluation metrics follow the definitions by Wang et al. \cite{Wang.2024} and include the relative pose error (RPE), relative rotation error (RRE), and absolute pose error (APE).

Table~\ref{tab:snail} and Table \ref{tab:hercules} show the results of the public benchmark. Across both datasets, our algorithm demonstrates superior performance in low-speed maneuvers, such as the \textit{113/1} and \textit{123/3} sequences from the SNAIL dataset and the \textit{Parking Lot Night 3} and \textit{Street Day 1} sequences from the Hercules dataset. In these cases, where velocities remain below 10 m/s, our method achieves the lowest RPE and RRE, along with competitive APE values, even in the absence of loop closure.
Our algorithm achieves an average RPE of 0.06 m on these low-speed sequences, outperforming the next best method, Doppler-SLAM, which averages 0.14 m. For higher-speed sequences ($v\geq$ 10 m/s), it maintains competitive performance with an average RPE of 0.23 m compared to 0.19 m for Doppler-SLAM.

The proposed algorithm performs particularly well during low-speed maneuvers such as parking, primarily due to the tight coupling of radar-Doppler measurements within the estimation framework. This coupling increases the bearing accuracy of detected features, thereby improving the information content of Doppler updates. In addition, the 4D outlier rejection and data association enables effective clutter suppression, further increasing estimation accuracy. Our approach is especially beneficial in parking scenarios, where features remain within the vehicle's FOV for an extended period, allowing each feature to contribute more information to the filter. At higher speeds, however, features exit the FOV rapidly, reducing their individual information contribution.

\section{Conclusions}
\label{sec:conclusions}
We developed a radar-inertial SLAM algorithm, that performs especially well at low vehicle speeds ($v<10$ m/s). Using a proprietary dataset, we showed that the performance is reached consistently using statistical evaluation methods on a large dataset. Our algorithm allows for compact measurement updates, tightly coupling Doppler, bearing and range measurements, while the information-based pruning and matching strategy ensures robust correspondences and consistent filter health. Last, we provided an extension to multi-radar systems that allows for cross-matching of features, which is especially beneficial for parking maneuvers, where features quickly enter and leave a radar sensor's FOV. Using publicly available datasets, we benchmarked our approach against state-of-the-art radar SLAM approaches and could show increased performance at low vehicle speeds ($v<10$ m/s). At higher speeds, our algorithm still produces competitive results.
Future work should focus on the integration and joint calibration of wheel odometry, inertial measurements, and vehicle models. In doing so, wheel-odometry-based approaches could be further refined to achieve the centimeter-level accuracy required for automated parking, while also benefiting from the advantages of radar-based SLAM.

\bibliographystyle{IEEEtran}
\bibliography{bib/literatur}\vspace{-1cm}

\end{document}